\documentclass[10pt,twocolumn,letterpaper]{article}

\usepackage[pagenumbers]{cvpr} % arXiv version (non-anonymous, with page numbers)

\definecolor{cvprblue}{rgb}{0.21,0.49,0.74}
\usepackage[breaklinks,colorlinks,allcolors=cvprblue]{hyperref}
\usepackage{algorithm}
\usepackage{algorithmic}
\usepackage{amssymb}
\usepackage{algorithm}            % For the algorithm float

\usepackage{amsmath}
\usepackage{booktabs}
\usepackage{multirow}
\usepackage{amsmath,amssymb}
\usepackage{array}
\usepackage{xcolor} 
\usepackage[table]{xcolor}

\def\paperID{18119} % *** Enter the Paper ID here
\def\confName{CVPR}
\def\confYear{2026}

\title{When Local Semantics Disagree: Learning to Detect Fake News \\ through Weakly Supervised Cross-Modal Reasoning}

\author{
Zeheng Wang\\
Northeast Normal University\\
{\tt\small wangzeheng624@nenu.edu.cn}
\and
Hao Zhang\\
University of the Chinese Academy of Sciences
\and
Yige Wang\\
Boston University
\and
Xi Xiao\\
University of Alabama at Birmingham
\and
Hao Xu\\
Harvard University\\
{\tt\small haxu@bwh.harvard.edu}
}

\begin{document}
\maketitle
\begin{abstract}
The rapid proliferation of multimodal social media content has made fake news detection a pressing challenge for the vision-language community.  
Most existing approaches rely on coarse global similarity between text and images, which overlooks subtle, uneven cross-modal inconsistencies—such as sentiment-scene mismatches, object-level manipulations, or semantic dissonance between visual and textual cues.  
Moreover, these models treat all inconsistencies as equally informative, ignoring domain-specific variation in how such conflicts manifest, which limits both generalization and interpretability.  
To address these challenges, we propose \textbf{DMIL}, a dual-branch multiple instance learning framework that formulates multimodal fake news detection as a weakly supervised learning problem.  
Each news article is treated as a bag of visual and textual instances processed through two complementary branches: an embedding-space branch that aggregates global semantics via expert-guided modality fusion, and an instance-space branch that models cross-modal inconsistencies across local-to-global granularities.  
A retrieval-augmented, domain-adaptive router further contextualizes inconsistencies using retrieved domain exemplars and dynamically fuses multi-level cues.  
Extensive experiments on public misinformation benchmarks demonstrate that DMIL achieves competitive state-of-the-art performance and demonstrates strong discriminative capability against fake news across different domains.
\end{abstract}
\section{Introduction}
% \begin{figure}[htbp]
%     \centering
%     \includegraphics[width=1\linewidth]{figure/intro.png}
%     \caption{Fake and real news. \textcolor{red}{just 1 row (maybe 1 real + 2 fake ) would be enough, and provide domain label for each of these examples! And I would prefer you can use inconsistency score to label key conflicts and show the overall logits for real and fake in the results section.  }}
%     \label{fig:MIL_types}
% \end{figure}

\begin{figure}[htbp]
    \centering
    \includegraphics[width=1\linewidth]{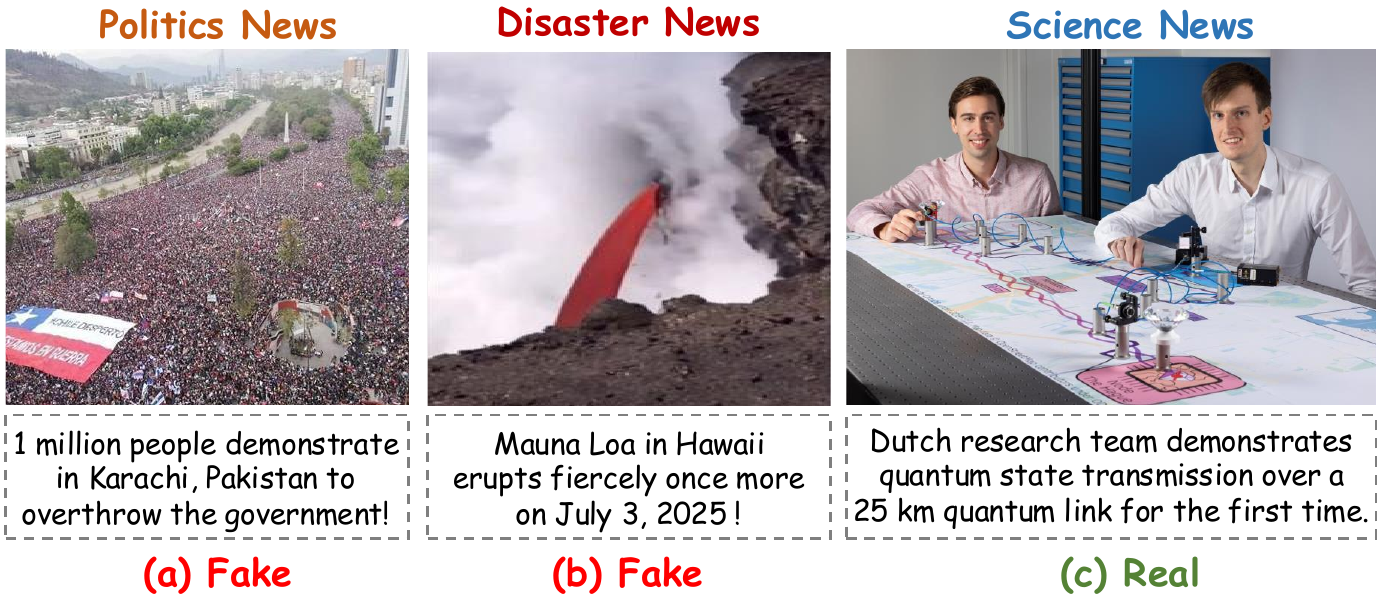}
%     \caption{ In Figure 1, the political news (a) demonstrates text-image inconsistency where the visible Chilean flag contradicts the description of a Pakistani coup, confirming it as fake. The disaster news (b), however, falsely pairs text about Mauna Loa’s eruption with an image of a different volcano, violating this field’s strict image-text alignment standards. Conversely, the science news (c) uses a conceptual illustration rather than depicting the actual 25km quantum link described, a discrepancy not considered fake in this domain since such visuals aid public understanding of abstract concepts. News items (a) and (b) are fake news, whereas (c) is real news.
% stract concepts.}
\caption{Examples illustrating domain-dependent tolerance for image-text inconsistency. 
(a) Political news shows the Chilean flag while describing a Pakistani coup, indicating a cross-modal inconsistency and labeling it as fake. 
(b) Disaster news misrepresents Mauna Loa’s eruption using an image of a different volcano, violating domain alignment norms and thus marked as fake. 
(c) Science news employs a conceptual illustration of a 25 km quantum link, which is acceptable for public communication in this domain and not considered fake.}

    \label{fig:MIL_types}
\end{figure}

The rapid dissemination of misinformation on social media has intensified the demand for automated fake news detection systems that can scale beyond manual fact-checking~\cite{shu2017fakenews,zhou2020survey,shu2020fakenewsnet,shu2020leveraging}. As online posts increasingly combine text and images, \textbf{multimodal fake news detection (MFND)} has become essential for assessing credibility in real-world scenarios~\cite{wu2024ueei,yu2022ddchms}. Early multimodal models~\cite{wang2018eann,khattar2019mvae,singhal2020spotfake} fuse global visual and textual embeddings to predict article veracity. However, such global fusion implicitly assumes that all regions and tokens are equally relevant, making these models insensitive to subtle, uneven cross-modal inconsistencies---for instance, a mismatch between a caption’s sentiment and an image’s visual tone.

Recent \textit{inconsistency-centric} approaches~\cite{tan2020inconsistency,xiong2023trimoon,wu2023mfir,guo2025crossmodal} have sought to detect semantic conflicts between modalities by aligning image regions and textual entities. While these methods improve interpretability, they still rely on {strong supervision}: each article is labeled as real or fake, but instance-level cues (e.g., which image region or sentence causes inconsistency) remain unannotated. Consequently, the model faces uncertainty about which multimodal pairs actually determine misinformation. Yet none of these methods explicitly adopt a weakly supervised formulation to address this uncertainty.

\begin{figure*}[htbp]
  \centering
  \includegraphics[width=0.9\textwidth]{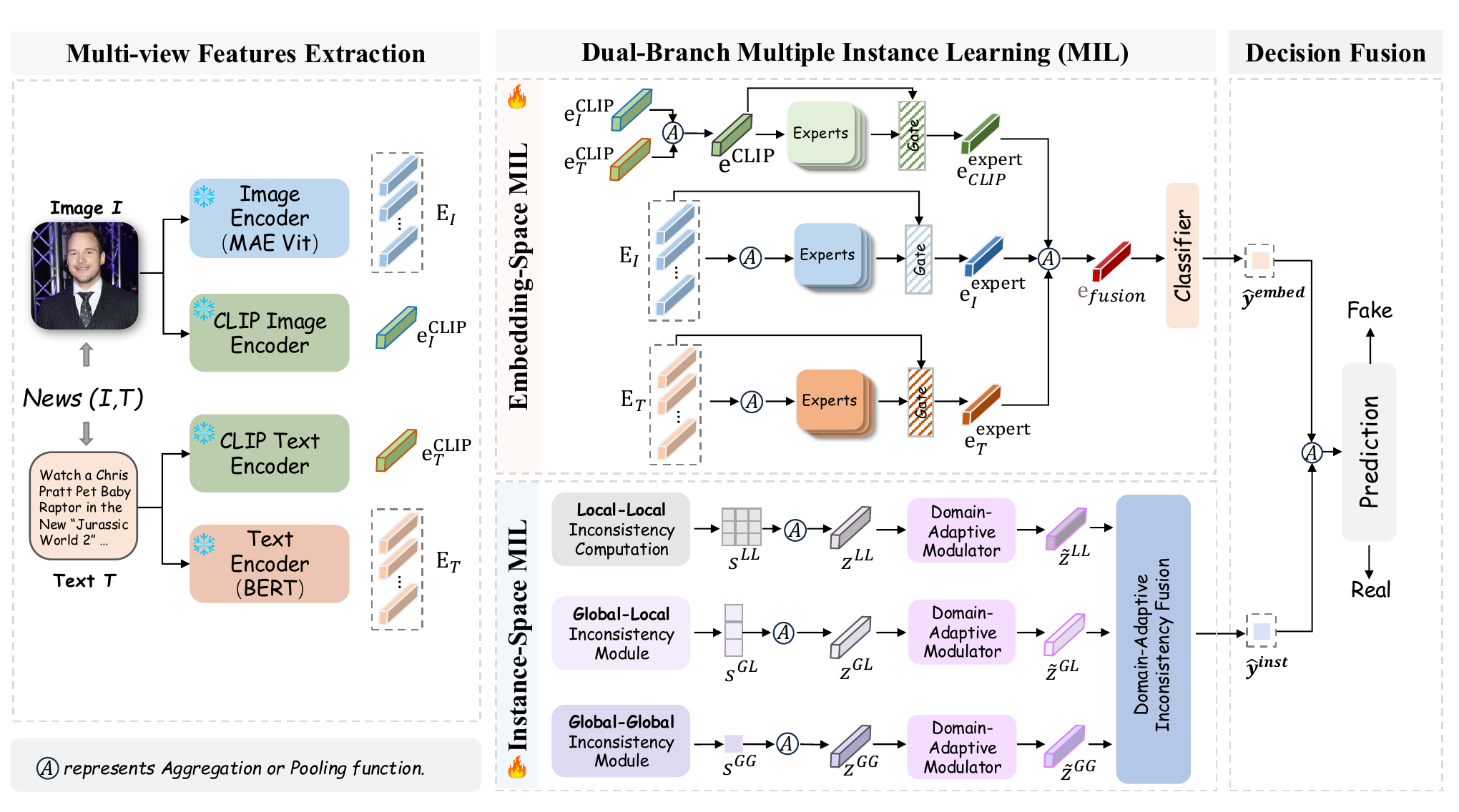}
  \caption{Overview of the proposed \textbf{DMIL} framework. Each article is modeled as a bag of multimodal instances processed by two complementary branches: an embedding-space MIL branch for global semantic reasoning, and an instance-space MIL branch for fine-grained inconsistency modeling, guided by domain-aware adaptation.}
  \label{fig:wide-image}
\end{figure*}

\textbf{Why weak supervision?}  
In multimodal misinformation data, article-level labels are available, but instance-level supervision is absent. The model must therefore infer which image–text combinations contribute to the label. This naturally fits the \textit{Multiple Instance Learning (MIL)} framework~\cite{maron1997framework,foulds2010review}, which treats each article as a \textit{bag} of multimodal instances (e.g., text–region pairs) and learns to predict bag-level veracity from ambiguous instance contributions. MIL provides a principled mechanism for weak supervision, mitigating noisy alignment while preserving interpretability.

\textbf{Why domain awareness?}  
Cross-modal inconsistencies manifest differently across domains: emotional–scene incongruities may flag misinformation in social or political news but appear benign in entertainment. Prior models treat all domains uniformly~\cite{nan2021mdfend,silva2021cdmmfn}, limiting adaptability and generalization. Incorporating domain-aware reasoning enables the detector to modulate the strength and meaning of inconsistencies according to contextual priors.

To address the above challenges, we propose {Dual-Branch Multiple Instance Learning for Multimodal Fake News Detection} (\textbf{DMIL}), a unified framework that explicitly formulates fake news detection as a weakly supervised learning task under the Multiple Instance Learning (MIL) paradigm. Each news article is treated as a bag composed of multi-granularity visual and textual instances. Our framework employs two complementary MIL branches: (1) an embedding-space branch, which performs progressive feature aggregation and fusion across textual, visual, and cross-modal representations to learn a unified global embedding optimized for bag-level veracity prediction; and (2) an instance-space branch, which explicitly computes multi-granularity cross-modal inconsistencies across three levels (local-local, global-local, and global-global) to support interaction-aware predictions. To improve adaptability across diverse news contexts, we introduce a \textit{domain-aware reasoning mechanism} that modulates each inconsistency type using a multi-view domain representation and aggregates them via a router network. By leveraging both global semantic coherence and localized mismatches, DMIL achieves robust, interpretable, and domain-sensitive veracity prediction. In summary, our key contributions are: 
\begin{itemize}
    \item \textbf{Weakly supervised MIL formulation.} We explicitly cast multimodal fake news detection as a MIL problem, learning from article-level labels while discovering latent instance-level cues.
    \item \textbf{Dual-branch MIL architecture.} DMIL integrates two complementary perspectives: (i) an \textit{embedding-space branch} that aggregates modality-specific representations through expert-guided fusion, and (ii) an \textit{instance-space branch} that computes local-local, global-local, and global-global inconsistency scores for interpretable reasoning.
    \item \textbf{Domain-aware reasoning mechanism.} A retrieval-augmented hyper-network contextualizes inconsistencies using both explicit domain labels and semantically similar retrieved posts, enabling fine-grained, context-aware adaptation.
    \item \textbf{Empirical effectiveness.} Extensive experiments on Weibo, Weibo-21, and GossipCop demonstrate that DMIL achieves state-of-the-art performance, particularly in cross-domain settings, while maintaining interpretability through structured inconsistency modeling.
\end{itemize}

\section{Related Work}

This section reviews prior work along three dimensions: general multimodal fake news detection, inconsistency-centric methods, and domain-adaptive approaches.

\subsection{General Multimodal Fake News Detection}
Early MFND models primarily focused on global feature fusion to integrate text and image modalities. attRNN~\cite{jin2017attRNN} combined visual attention with LSTM-based text modeling, MVAE~\cite{khattar2019mvae} employed a multimodal variational autoencoder, and SpotFake~\cite{singhal2020spotfake} fused XLNet and ResNet embeddings. SAFE~\cite{zhou2020safe} introduced a similarity-aware fusion mechanism, MCAN~\cite{xue2021mcnn} applied co-attention between modalities, and COOLANT~\cite{wang2023coolant} leveraged contrastive learning for alignment. Later works such as FND-CLIP~\cite{zhou2023clip} and MSACA~\cite{wang2024msaca} utilized pretrained vision–language models to enhance semantic coherence. However, all these methods operate at the holistic level and do not explicitly identify localized semantic contradictions. More broadly, recent progress in synthetic and forged content detection~\cite{tan2026veritas,tan2026videoveritas,shi2026hydraprompt,ma2026cm,ma2026clip,ma2026flexible}, security of generative vision-language models~\cite{DBLP:conf/iclr/PanCWDW26,DBLP:journals/corr/abs-2605-07399}, multimodal affective understanding and reasoning~\cite{zhu2025emosym,zhu2026uniemo,11658652,wang2026affectagent,wang2026navigating,liu2026aullm++,wang2026micro,10.1145/3767308.3835254,10.1007/978-981-92-4529-1_3}, vision-language-action modeling~\cite{zhu2026h,zhu2026delta,zhu2026ati,zhu2026learning}, multimodal large language models for change captioning~\cite{wang2026rsiccllm,wang2026dfm}, signed-graph structural modeling~\cite{gao2026polarityasymmetricstructuralcalibrationlink,wang2026signedgraph}, and speech synthesis~\cite{wang2027manchutts} further demonstrates the value of fine-grained cross-modal reasoning.

\subsection{Inconsistency-Centric Multimodal Fake News Detection}
To improve interpretability, inconsistency-centric methods explicitly capture modality conflicts. DIDAN~\cite{tan2020inconsistency} computed global mismatch scores between visual and textual features; TRIMOON~\cite{xiong2023trimoon} refined fusion through two-stage inconsistency filtering; MFIR~\cite{wu2023mfir} introduced an inconsistency reasoning module for explainable predictions; and CH-Match~\cite{guo2025crossmodal} modeled consistency and heterogeneity jointly. Despite their effectiveness, these methods rely on full supervision and lack a principled weakly supervised learning mechanism such as MIL. Moreover, they often treat all domains identically, ignoring the domain-dependent nature of cross-modal conflicts.

\subsection{Domain-Aware and Cross-Domain MFND}
Cross-domain generalization has been explored through domain embeddings or disentangled representations. MDFEND~\cite{nan2021mdfend} employed domain-specific experts with gating, CDMFND~\cite{silva2021cdmmfn} combined textual and social signals, and MMDFND~\cite{tong2024mmdfnd} disentangled shared versus private features. DAMMFND~\cite{lu2025dammfnd} introduced domain-enhanced discriminators for better cross-domain robustness. Nevertheless, these methods treat domain information as a single categorical input and do not integrate fine-grained contextual retrieval to guide adaptation. 

In contrast, our DMIL framework unifies these perspectives: it formulates fake news detection as a weakly supervised MIL problem, models multi-granularity inconsistencies, and introduces a retrieval-augmented domain-aware mechanism to dynamically contextualize inconsistency signals.

\section{Background}
\subsection{Multiple Instance Learning}

Multiple Instance Learning (MIL) is a form of weakly supervised learning in which only \textit{bag-level} labels are available, while the labels of individual \textit{instances} within each bag remain unknown~\cite{maron1997framework,herrera2016multiple,carbonneau2018multiple}. This paradigm is especially useful when fine-grained annotations are expensive or impractical to obtain. For example, in weakly supervised object detection, an image may be labeled as containing an object without specifying its precise location, allowing the model to infer which regions correspond to the target~\cite{quellec2017multiple}.

\begin{figure}[htbp]
    \centering
    \includegraphics[width=1\linewidth]{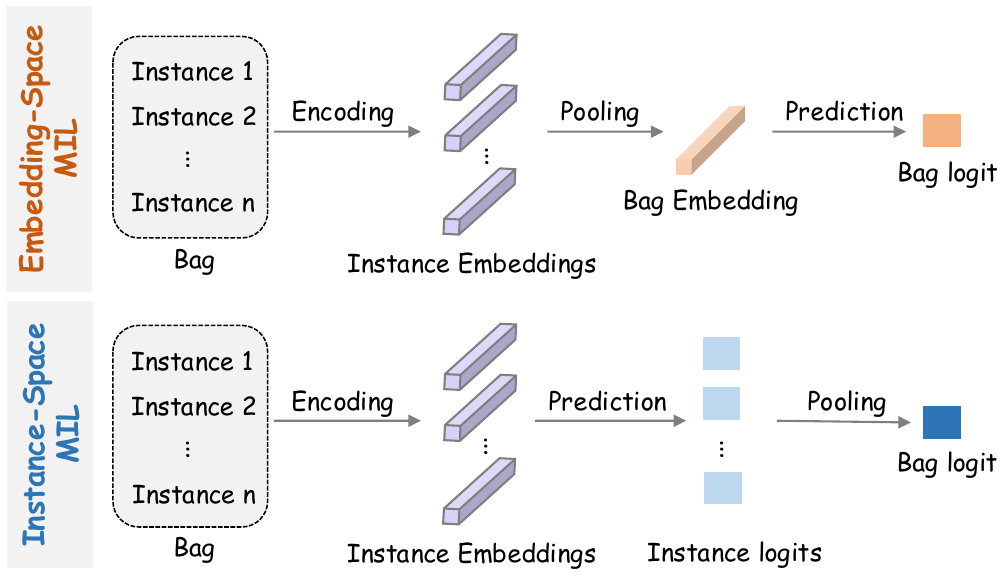}
    \caption{Illustration of two MIL paradigms: Embedding-Space MIL and Instance-Space MIL.}
    \label{fig:MIL_types}
\end{figure}

MIL frameworks typically operate under two main assumptions \cite{foulds2010review}: the standard assumption, where a bag is positive if any instance is positive and negative otherwise; and the collective assumption, where the bag label depends on the joint contribution of multiple instances.
The collective assumption better suits high-level reasoning tasks—such as multimodal fake news detection—where the overall veracity of a news article depends on the collective consistency of its textual and visual components.
Under this setting, two primary architectural paradigms exist:
\begin{itemize}
    \item \textbf{Embedding-Space MIL:} Each instance is first encoded into a feature embedding, which is then aggregated to form a bag-level representation for classification.
    \item \textbf{Instance-Space MIL:} Instance-level scores or logits are first computed and subsequently pooled to infer the bag label.
\end{itemize}
The former follows an \textit{Encoding → Pooling → Prediction} flow, while the latter adopts an \textit{Encoding → Prediction → Pooling} flow, providing finer interpretability through instance-level reasoning.

\subsection{Problem Formulation: Multi-Granularity MIL for Fake News Detection}

We formulate multimodal fake news detection as a MIL problem under the \textit{collective assumption}, where each news article is treated as a bag containing multiple cross-modal instances at varying granularities.

Let \( (I, T) \) represent the multimodal input, where  
\( I = \{ p_1, p_2, \dots, p_{L_I} \} \) denotes the set of image patches and  
\( T = \{ t_1, t_2, \dots, t_{L_T} \} \) denotes the sequence of text tokens.  
The goal is to predict the veracity label \( y \in \{0, 1\} \) for each bag without explicit supervision on the patch–token pairings responsible for misinformation.

\paragraph{Embedding-Space MIL.}
In the embedding-space branch, modality-specific instances are first mapped to embedding vectors via an encoder function \( \phi(\cdot) \):
\begin{equation}
X^{\text{emb}} = \{ \phi(x_i) \mid x_i \in I \cup T \}.
\end{equation}
A pooling operator aggregates these embeddings to obtain a global bag representation \( e_{\text{fused}} \), which is then classified by a function \( f(\cdot) \):
\begin{equation}
\mathbb{P}_{\text{emb}}(y \mid X^{\text{emb}}) = f(e_{\text{fused}}).
\end{equation}
This branch captures holistic, global-level semantics through expert-guided fusion across visual and textual modalities.

\paragraph{Instance-Space MIL.}
In the instance-space branch, each news article is modeled as a bag of cross-modal interaction pairs covering multiple granularities:
\begin{equation}
X^{\text{inst}} = \{ (p,t),\; (p,T),\; (I,t),\; (I,T) \},
\end{equation}
where \( p \in I \) denotes an image patch and \( t \in T \) a text token.  
These interaction types correspond to local-local, global-local, and global-global alignments between visual and textual features.

Each instance \( x_i \in X^{\text{inst}} \) is assigned an inconsistency score \( h(x_i) \), and the bag-level prediction is obtained via a permutation-invariant pooling function \( g(\cdot) \):
\begin{equation}
\mathbb{P}_{\text{inst}}(y \mid X^{\text{inst}}) = g\left( \{ h(x_i) \mid x_i \in X^{\text{inst}} \} \right).
\end{equation}
This formulation enables the model to reason over fine-grained cross-modal inconsistencies while maintaining weak supervision at the article level.

\paragraph{Motivation for Dual-Branch MIL.}
The embedding-space and instance-space branches provide complementary perspectives: the former captures aggregated global semantics, while the latter identifies localized semantic conflicts. By integrating both within a unified framework, our Dual-Branch MIL (DMIL) achieves interpretable and domain-adaptive fake news detection under weak supervision.

\section{Method}

We propose the \textbf{Dual-branch Multiple Instance Learning (DMIL)} framework for multimodal fake news detection (MFND).  
DMIL unifies two complementary reasoning branches—\textit{Embedding-Space MIL} and \textit{Instance-Space MIL}—within a coherent weakly supervised formulation.  
As illustrated in Figure~\ref{fig:wide-image} and Algorithm~\ref{alg:dmil}, the model operates in three sequential phases:
\textit{(i) Multi-view feature extraction}, 
\textit{(ii) Dual-branch multiple instance learning}, and
\textit{(iii) Dual-branch decision fusion}.
This design integrates global semantic coherence with fine-grained cross-modal inconsistency reasoning. 

\begin{algorithm}[t]
\caption{Dual-Branch Multiple Instance Learning (DMIL) for Multimodal Fake News Detection}
\label{alg:dmil}
\begin{algorithmic}[1]
\REQUIRE Text $T$, image $I$, domain $d$, retrieval memory $\mathcal{B}$
\ENSURE Predicted veracity $\hat{y}$

\STATE \textbf{Stage 1: Multi-View Feature Extraction.}\\
Obtain multimodal embeddings:  
textual embeddings $\mathbf{E}_T$,  
visual embeddings $\mathbf{E}_I$, 
and cross-modal global embeddings $\mathbf{e}_T^{\text{CLIP}}, \mathbf{e}_I^{\text{CLIP}}$.

\STATE \textbf{Stage 2a: Embedding-Space MIL Branch.}
\\
Aggregate modality embeddings via attention pooling to obtain  
$\mathbf{e}_T^{\text{agg}}$ and $\mathbf{e}_I^{\text{agg}}$;  
refine with expert modules to produce $\mathbf{e}_m^{\text{expert}}$ ($m\!\in\!\{\text{T},\text{I},\text{CLIP}\}$);  
compute modality weights $\boldsymbol{\alpha}$ via a Softmax gate;  
fuse weighted representations into $\mathbf{e}_{\text{fusion}}$;  
predict embedding-level score $\hat{y}^{\text{emb}}$.

\STATE \textbf{Stage 2b: Instance-Space MIL Branch.}
\\
Construct cross-modal instance set $X^{\text{inst}}=\{(p,t),(p,T),(I,t),(I,T)\}$;  
compute inconsistency vectors $z^{\text{LL}}$, $z^{\text{GL}}$, $z^{\text{GG}}$;  
adapt them using retrieval-augmented domain modulation with  
parameters $(\boldsymbol{\gamma},\boldsymbol{\beta})$ and obtain $\tilde{z}^m$;  
aggregate modulated features with learned weights $w_m$  
to derive fused representation $z^{\text{fused}}$;  
predict instance-level score $\hat{y}^{\text{inst}}$.

\STATE \textbf{Stage 3: Dual-Branch Decision Fusion.}
\\
Fuse embedding- and instance-level predictions to obtain  
$\hat{y} = \sigma(\mathrm{MLP}([\hat{y}^{\text{inst}}\Vert\hat{y}^{\text{emb}}]))$.

\STATE \textbf{Optimization.}
\\
Train the model with binary cross-entropy loss: 
$
\mathcal{L} = -y\log(\hat{y}) - (1 - y)\log(1 - \hat{y})
$
\end{algorithmic}
\end{algorithm}

\subsection{Multi-View Feature Extraction}

To capture complementary visual, textual, and cross-modal cues, we employ three encoders:  
a BERT-based text encoder, a MAE-based image encoder, and a CLIP-based multimodal encoder.  
The resulting representations serve as the foundational inputs for both MIL branches.

\subsubsection{Textual Token Embedding}
The text \(T = \{ t_1, t_2, \dots, t_{L_T} \}\) is tokenized and passed through a pre-trained BERT model~\cite{devlin2019bert} to obtain contextual token embeddings:
\begin{equation}
\mathbf{E}_T = \{\, \mathbf{e}^{(j)}_T \}_{j=1}^{L_T} \in \mathbb{R}^{L_T \times D_T},
\end{equation}
where \(L_T\) is the number of tokens and \(D_T\) is the embedding dimension of BERT.

\subsubsection{Visual Patch Embedding}
The image \(I = \{ p_1, p_2, \dots, p_{L_I} \}\) is divided into patches and encoded by a pre-trained Masked Autoencoder (MAE)~\cite{he2022mae}, yielding:
\begin{equation}
\mathbf{E}_I = \{\, \mathbf{e}^{(p)}_I \}_{p=1}^{L_I} \in \mathbb{R}^{L_I \times D_I},
\end{equation}
where \(L_I\) is the number of patches and \(D_I\) is the embedding dimension of the MAE.

\subsubsection{Cross-Modal Aligned Global Embedding}
We adopt CLIP~\cite{a2021clip} to obtain global image–text embeddings aligned in a shared semantic space:
\begin{equation}
\mathbf{e}_T^{\text{CLIP}},\ \mathbf{e}_I^{\text{CLIP}} \in \mathbb{R}^{D_{\text{CLIP}}}.
\end{equation}
These embeddings provide coarse global semantics, while the token and patch features retain local granularity.

\subsection{Embedding-Space MIL}

In this branch, each article is modeled as a bag of multimodal embeddings:
\begin{equation}
X^{\text{emb}} = \{ \mathbf{E}_T,\, \mathbf{E}_I,\, \mathbf{e}_T^{\text{CLIP}},\, \mathbf{e}_I^{\text{CLIP}} \}.
\end{equation}

We first perform \textit{intra-modality aggregation} using attention pooling to obtain global text and image representations:
\begin{equation}
\mathbf{e}_T^{\text{agg}} = \operatorname{AttnPool}(\mathbf{E}_T), 
\quad
\mathbf{e}_I^{\text{agg}} = \operatorname{AttnPool}(\mathbf{E}_I).
\end{equation}

Next, we derive a unified cross-modal embedding from CLIP:
\begin{equation}
\mathbf{e}_{\text{CLIP}} = 
\operatorname{MLP}\!\left([\mathbf{e}_T^{\text{CLIP}} \,\Vert\, \mathbf{e}_I^{\text{CLIP}}]\right).
\end{equation}

Each aggregated modality representation is then refined via a modality-specific expert network gated by learned coefficients:
\begin{equation}
\mathbf{e}_m^{\text{expert}} = 
\sum_j \operatorname{Gate}_{m,j}(\mathbf{e}_m) 
\cdot \operatorname{Expert}_{m,j}(\mathbf{e}_m),
\end{equation}
where \(m \in \{\text{T}, \text{I}, \text{CLIP}\}\).  
A cross-modal weighting module computes normalized modality weights:
\begin{equation}
\begin{aligned}
\boldsymbol{\alpha} &= \mathrm{Softmax} \left( \mathrm{MLP} \left( 
[\mathbf{e}_{I}^{\text{expert}} \,\|\, \mathbf{e}_{T}^{\text{expert}} \,\|\, \mathbf{e}_{CLIP}^{\text{expert}}] \right) \right) \\
&= [\alpha_{\text{I}}, \alpha_{\text{T}}, \alpha_{\text{CLIP}}]
\end{aligned}
\end{equation}

The fused embedding is obtained as a weighted sum followed by a fully connected projection:
\begin{equation}
\mathbf{e}_{\text{fusion}} =
\operatorname{FC}\!\left(
\alpha_{\text{I}} \mathbf{e}_{\text{I}}^{\text{expert}}
+ \alpha_{\text{T}} \mathbf{e}_{\text{T}}^{\text{expert}}
+ \alpha_{\text{CLIP}} \mathbf{e}_{\text{CLIP}}^{\text{expert}}
\right).
\end{equation}

Finally, the bag-level prediction for this branch is produced via:
\begin{equation}
\hat{y}^{\text{emb}} = \operatorname{MLP}(\mathbf{e}_{\text{fusion}}).
\end{equation}

\subsection{Instance-Space MIL}

In the instance-space branch, each article is modeled as a bag of cross-modal interaction pairs covering multiple granularities:
\begin{equation}
X^{\text{inst}} = \{ (p,t),\ (p,T),\ (I,t),\ (I,T) \},
\end{equation}
corresponding respectively to local–local, local–global, global–local, and global–global interactions.  
Each instance yields an inconsistency score \( s^m \), with \(m\in\{\text{LL},\text{GL},\text{GG}\}\).

\subsubsection{Local-Local Inconsistency}
Token–patch inconsistency captures fine-grained semantic mismatches.  
Let
\[
Q = W_Q^T \mathbf{E}_T,\quad
K = W_K^I \mathbf{E}_I,\quad
V = W_V^I \mathbf{E}_I,
\]
where \(W_Q^T \in \mathbb{R}^{D \times D_T}\),
\(W_K^I, W_V^I \in \mathbb{R}^{D \times D_I}\),
and \(d_k = D / h\).  
Scaled dot-product attention computes token–patch affinities:
\begin{equation}
A^{\text{LL}} = \operatorname{softmax}\!\left(\frac{QK^\top}{\sqrt{d_k}}\right).
\end{equation}
The corresponding inconsistency matrix is its complement:
\begin{equation}
s^{\text{LL}} = \mathbf{1}_{L_T \times L_I} - A^{\text{LL}}.
\end{equation}
The vectorized representation is:
\begin{equation}
z^{\text{LL}} = 
\operatorname{SiLU}\!\left(
W^{\text{LL}} 
\frac{1}{L_T} \sum_{j=1}^{L_T} (s^{\text{LL}} V)_{j,:}
\right) \in \mathbb{R}^{D_{\text{feat}}}.
\end{equation}

\subsubsection{Global–Local Inconsistency}
Global–local inconsistency measures how well a global embedding aligns with local elements across modalities.  
For concreteness, we consider global text versus local image patches:
\[
Q = W_Q^{T\rightarrow I}\mathbf{e}_T^{\text{CLIP}}, \quad
K = W_K^{I\rightarrow T}\mathbf{E}_I, \quad
V = W_V^{I\rightarrow T}\mathbf{E}_I.
\]
The inconsistency vector is:
\begin{equation}
s^{\text{GL}} = \mathbf{1}_{1\times L_I} -
\operatorname{softmax}\!\left(\frac{QK^\top}{\sqrt{d_k}}\right),
\end{equation}
and its projected representation:
\begin{equation}
z^{\text{GL}} = 
\operatorname{SiLU}\!\left(
W^{\text{GL}}(s^{\text{GL}}\odot V)
\right)
\in \mathbb{R}^{D_{\text{feat}}}.
\end{equation}

\subsubsection{Global-Global Inconsistency}
At the coarsest level, we compare the global CLIP embeddings:
\[
\mathbf{e}_I^{\text{CLIP}},\ \mathbf{e}_T^{\text{CLIP}} \in \mathbb{R}^{D_{\text{CLIP}}}.
\]
Their squared Euclidean distance is:
\begin{equation}
d = \|\mathbf{e}_I^{\text{CLIP}} - \mathbf{e}_T^{\text{CLIP}}\|_2^2,
\end{equation}
and the RBF-transformed inconsistency score is:
\begin{equation}
s^{\text{GG}} = 2\left(1 - \exp\!\left(-\frac{d}{2\sigma^2}\right)\right),
\end{equation}
where \(\sigma\) controls sensitivity.  
The vectorized representation is:
\begin{equation}
z^{\text{GG}} =
W^{\text{GG}}\,\operatorname{SiLU}(W_I^{\text{GG}} s^{\text{GG}}).
\end{equation}

\subsubsection{Retrieval-Augmented Domain-Adaptive Mechanism}
\begin{figure}[htbp]
    \centering
    \includegraphics[width=1\linewidth]{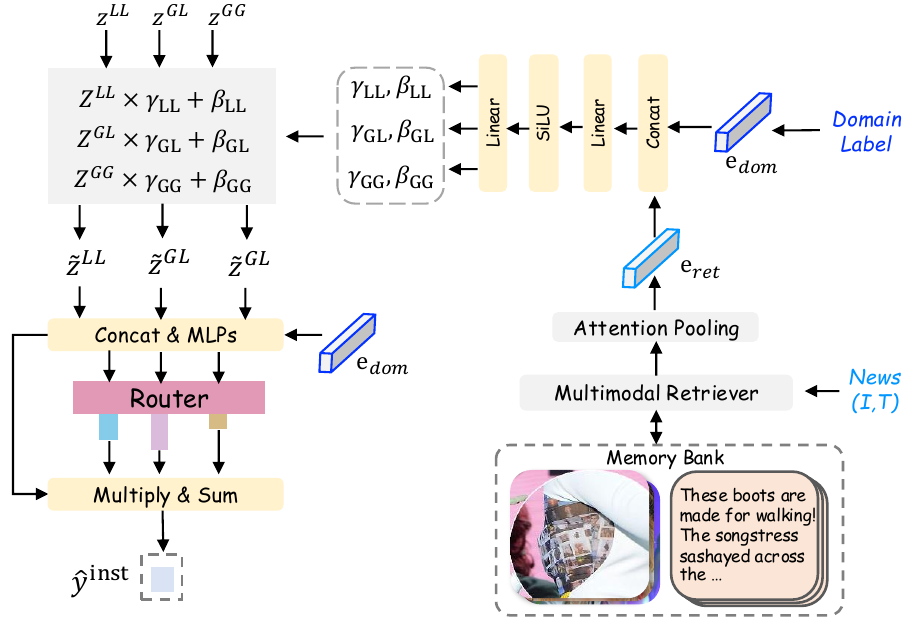}
    \caption{Overview of the Domain-Adaptive Modulator with trieval-augmented domain-adaptive reasoning and Domain-Adaptive Inconsistency Fusion.}
    \label{fig:dr}
\end{figure}

To enhance domain robustness, we introduce a retrieval-augmented mechanism that conditions inconsistency representations on domain-aware context.

A retrieval memory bank
\(\mathcal{B}=\{(I_j,T_j)\}_{j=1}^{|\mathcal{B}|}\)
stores image–text pairs.  
For a query \((I,T)\), global embeddings
\(\mathbf{e}_i^I,\mathbf{e}_i^T\)
are compared with memory items
\(\mathbf{e}_j^I,\mathbf{e}_j^T\)
to compute similarity:
\begin{equation}
\kappa_{ij} = 
\gamma_I \cos(\mathbf{e}_i^I,\mathbf{e}_j^I)
+ \gamma_T \cos(\mathbf{e}_i^T,\mathbf{e}_j^T),
\end{equation}
where \(\gamma_I,\gamma_T>0\) and \(\gamma_I+\gamma_T=1\).  
The top-\(K\) most similar samples are retrieved and attentively pooled:
\begin{equation}
\mathbf{e}_{\text{ret}} = 
\operatorname{AttnPool}\!\left\{ g(I_{r_k}, T_{r_k}) \right\}_{k=1}^{K}.
\end{equation}

Concatenating the retrieved context with a domain embedding
\(\mathbf{e}_{\text{dom}}\)
produces the input to a hyper-network:
\begin{equation}
(\boldsymbol{\gamma}, \boldsymbol{\beta}) =
\operatorname{HyperNet}\!\bigl([\mathbf{e}_{\text{dom}} \Vert \mathbf{e}_{\text{ret}}]\bigr)
\in \mathbb{R}^{3\times D_{\text{feat}}},
\end{equation}
which generates FiLM modulation parameters to adapt the inconsistency vectors:
\begin{equation}
\tilde{z}^m = \gamma_m \odot z^m + \beta_m, 
\qquad m \in \{\text{LL},\text{GL},\text{GG}\}.
\end{equation}

A router learns adaptive fusion weights \(\mathbf{w}\in[0,1]^3\) with \(\sum_m w_m = 1\):
\begin{equation}
z^{\text{fused}} = \sum_m w_m\,\tilde{z}^m.
\end{equation}
The bag-level prediction from the instance-space branch is then:
\begin{equation}
\hat{y}^{\text{inst}} = \operatorname{MLP}(z^{\text{fused}}).
\end{equation}

\subsection{Dual-Branch Decision Fusion}

To obtain the final bag-level prediction, we concatenate the outputs from both branches and apply a lightweight MLP:
\begin{equation}
\mathbb{P}(y \mid I,T) =
\sigma\!\left(
\operatorname{MLP}\!\left([\hat{y}^{\text{inst}} \Vert \hat{y}^{\text{emb}}]\right)
\right),
\end{equation}
where \(\sigma(\cdot)\) denotes the sigmoid activation function.

The model is optimized using binary cross-entropy loss:
\begin{equation}
\mathcal{L} =
-\,y\log(\hat{y}) - (1-y)\log(1-\hat{y}),
\end{equation}
where \(y \in \{0,1\}\) is the ground-truth veracity label.  
This dual-branch fusion allows DMIL to jointly reason over global coherence and fine-grained inconsistencies, achieving robust and interpretable weakly supervised detection.

\section{Experiments}

\begin{table*}[t]
\centering

\renewcommand{\arraystretch}{1.0}
\setlength{\tabcolsep}{6pt}
\footnotesize

\begin{tabular}{cccccccccccccc}

\toprule
\multirow{2}{*}{\textbf{Datasets}} &
\multirow{2}{*}{\textbf{Method}}  &
\multirow{2}{*}{\textbf{Sci.}}    &
\multirow{2}{*}{\textbf{Mil.}}    &
\multirow{2}{*}{\textbf{Edu.}}    &
\multirow{2}{*}{\textbf{Soc.}}    &
\multirow{2}{*}{\textbf{Pol.}}    &
\multirow{2}{*}{\textbf{Hlth.}}   &
\multirow{2}{*}{\textbf{Fin.}}    &
\multirow{2}{*}{\textbf{Ent.}}    &
\multirow{2}{*}{\textbf{Dis./Int}} &
\multicolumn{3}{c}{\textbf{Overall}} \\
\cmidrule(lr){12-14}
 & & & & & & & & & & &
 \textbf{F1} & \textbf{Acc} & \textbf{AUC} \\
\midrule
\multirow{8}{*}{Weibo}
 & MMoE$^{*}$        & 57.8 & \underline{91.1} & 85.1 & 88.5 & 73.5 & 82.6 & 81.3 & 83.0 & 88.3 & 87.4 & 87.4 & 95.0 \\
 & MoSE$^{*}$        & 79.3 & 73.8 & 83.4 & 91.2 & 76.4 & 85.9 & 79.1 & 84.4 & 88.3 & 89.0 & 89.0 & 95.4 \\
 & MDFEND$^{*}$      & 77.4 & \underline{91.1} & 89.7 & 90.2 & 76.3 & 87.8 & 80.8 & 88.1 & 87.4 & 90.4 & 90.4 & 96.5 \\
 & M$^{3}$FEND$^{*}$ & 79.2 & 90.3 & {92.3} & 91.2 & 76.5 & 86.3 & {89.9} & {89.9} & 87.6 & 92.8 & 92.8 & 96.9 \\
 & KATMF             & {83.1} & 90.8 & {92.4} & 89.5 & \underline{82.3} & 89.8 & {90.3} & {90.4} & {89.4} & {92.9} & {93.0} & {96.9} \\
 & MMDFND$^{*}$      & 82.4 & {91.1} & {94.1} & {93.9} & 73.5 & {91.3} & {91.7} & {91.7} & 88.8 & {93.4} & {93.4} & {97.2} \\
 & DAMMFND$^{*}$     & \underline{85.3} & \underline{91.1} & \textbf{95.6} & \underline{94.3} & {82.2} & \underline{93.9} & \textbf{93.7} & \textbf{95.6} & \underline{92.8} & \underline{94.3} & \underline{94.4} & \textbf{98.3} \\
 & \textbf{DMIL}     & \textbf{90.3} & \textbf{92.4} & \underline{94.3} & \textbf{94.6} & \textbf{85.2} & \textbf{94.2} & \underline{92.1} & \underline{93.7} & \textbf{94.9} & \textbf{95.1} & \textbf{95.1} & \textbf{98.3} \\

\midrule
\multirow{8}{*}{Weibo-21}
 & MMoE$^{*}$        & {87.5} & {91.1} & 87.0 & 87.5 & 86.2 & {93.6} & 85.6 & 88.8 & 87.7 & 89.4 & 89.4 & 95.4 \\
 & MoSE$^{*}$        & 85.0 & 88.5 & 88.1 & 87.2 & 88.0 & 91.7 & 86.7 & 89.1 & 86.7 & 89.3 & 89.4 & 95.4 \\
 & MDFEND$^{*}$      & 83.0 & {93.8} & 89.1 & 89.8 & 88.0 & {94.0} & {89.5} & 90.6 & 90.0 & 91.3 & 91.3 & 97.0 \\
 & M$^{3}$FEND$^{*}$ & 82.9 & {95.0} & 89.9 & 90.8 & 88.2 & \underline{94.6} & 90.0 & 93.1 & 88.9 & 92.1 & 92.1 & 97.5 \\
 & KATMF             & 91.4 & 92.2 & {92.8} & 89.5 & 90.2 & 91.4 & 87.1 & 93.7 & {92.0} & 93.3 & 93.3 & {97.7} \\
 & MMDFND$^{*}$      & \underline{93.7} & \underline{95.3} & 85.2 & \underline{94.5} & {96.5} & 92.0 & 88.4 & {95.9} & 91.9 & {93.9} & {93.9} & {97.7} \\
 & DAMMFND$^{*}$     & {93.2} & \underline{95.3} & \underline{93.2} & 91.7 & \underline{98.2} & 91.9 & \underline{94.1} & \textbf{99.0} & \textbf{94.5} & \underline{94.7} & \underline{94.7} & \underline{98.5} \\
 & \textbf{DMIL}     & \textbf{94.1} & \textbf{95.6} & \textbf{96.7} & \textbf{99.7} & \textbf{98.5} & \textbf{95.2} & \textbf{94.6} & \underline{98.3} & \textbf{94.5} & \textbf{95.4} & \textbf{95.4} & \textbf{99.3} \\

\bottomrule
\end{tabular}
\caption{Performance comparison across individual domain categories and overall results (\%) on the Weibo and Weibo-21 datasets.
 {Bold} indicates the best result; {underline} indicates the second-best. $^{*}$denotes results reported from the original papers.}
\label{tab:detection-comparison}
\end{table*}

\subsection{Experimental Settings}
In this section, we present the experimental setup regarding dataset, baselines and elevation metrics. Additional implementation and training details are provided in the supplementary materials.
\subsubsection{Dataset}
We use three publicly available datasets: Weibo \cite{wang2018eann}, Weibo-21 \cite{nan2021mdfend}, and GossipCop \cite{shu2020fakenewsnet}. The Weibo dataset consists of 9,528 news articles (4,745 genuine and 4,783 fake). The Weibo-21 dataset consists of 9,127 news articles (4,640 genuine and 4,487 fake). For domain adaptation, the dataset was partitioned into nine topical domains—finance, health, military, science, politics, disasters, education, entertainment, and society—following standardized processing protocols aligned with established benchmarking practices.

% We use three publicly available datasets: Weibo \cite{wang2018eann}, Weibo-21 \cite{nan2021mdfend}, and GossipCop \cite{shu2020fakenewsnet}. The Weibo dataset includes 7,532 training samples (3,749 genuine and 3,783 fake news items) and 1,996 testing samples (996 genuine and 1,000 fake). For domain adaptation purposes, we divided this dataset into nine distinct topical domains: finance, health, military, science, politics, international affairs, education, entertainment, and society, following standardized processing guidelines. The Weibo-21 dataset consists of 9,127 news articles (4,640 genuine and 4,487 fake), which were similarly segmented into training and testing subsets based on established benchmarking procedures. 
% Additionally, GossipCop, sourced from a fact-checking platform, contains 10,010 training instances (7,974 real and 2,036 fake) and 2,830 testing instances (2,285 real and 545 fake). \textcolor{red}{GossipCop doesn't provide a label for domain, you need to mention how you got this.}

\subsubsection{Baselines} 
We select 7 representative baseline methods: MMoE \cite{ma2018mmoe}, KATMF \cite{khattar2019mvae}, MoSE \cite{qin2020mose}, MDFEND \cite{nan2021mdfend}, M3DFEND \cite{zhu2023m3fend},   MMDFND \cite{tong2024mmdfnd}, and DAMMFND \cite{lu2025dammfnd}.

%We select representative baseline methods: MMoE \cite{ma2018mmoe}, MoSE \cite{qin2020mose}, MDFEND \cite{nan2021mdfend}, and M3DFEND \cite{zhu2023m3fend}, FND-CLIP \cite{zhou2023clip}, MRHFR \cite{wu2023multireading}, CAFE \cite{chen2022cafe}, CMC \cite{wei2022cmd}, and BMR \cite{ying2023bmr}, KATMF \cite{khattar2019mvae}, MMDFND \cite{tong2024mmdfnd}, and DAMMFND \cite{lu2025dammfnd}.

%\textcolor{blue}{We don't have a baseline for the inconsistency category. Some baselines are not domain-aware, such as mmoe and mose} We select representative baseline methods from three distinct categories. Multi-domain methods include models such as MMoE \cite{ma2018mmoe}, MoSE \cite{qin2020mose}, MDFEND \cite{nan2021mdfend}, and M3DFEND \cite{zhu2023m3fend}. Multimodal methods encompass FND-CLIP \cite{zhou2023clip}, MRHFR \cite{wu2023multireading}, CAFE \cite{chen2022cafe}, CMC \cite{wei2022cmd}, and BMR \cite{ying2023bmr}. Finally, we consider hybrid multi-domain and multimodal approaches, specifically KATMF \cite{khattar2019mvae}, MMDFND \cite{tong2024mmdfnd}, and DAMMFND \cite{lu2025dammfnd}, to assess models capable of leveraging both cross-domain and cross-modal information.

\subsubsection{Evaluation Metrics}
We evaluate model performance using standard classification metrics: Accuracy, Precision, Recall, and F1-score. 
%Together, these metrics provide a comprehensive assessment of the model’s ability to correctly identify fake and real news instances.

\subsubsection{Implementation Details}
We adopt mae-pretrain-vit-base \cite{he2022mae} as the pretrained checkpoint for encoding images. Regarding text encoding, we apply the bert-base-chinese \cite{devlin2019bert} model to the Chinese datasets (Weibo and Weibo-21), while utilizing bert-base-uncased \cite{devlin2019bert} for the GossipCop dataset. To extract CLIP embeddings, clip-vit-base-patch16 \cite{a2021clip} is selected for GossipCop, and AltCLIP \cite{chen2023altclip} is chosen for Chinese datasets. Each Multi-Layer Perceptron (MLP) in our model comprises two linear layers interconnected by a Sigmoid Linear Unit (SiLU) activation. 

In our architecture, we adopt an Adaptive Multimodal Experts structure tailored to different modality streams. Specifically, we apply expert modules to textual, visual, and multimodal fused representations through a dynamic gating mechanism, independent of domain information. The expert networks used during training consist of convolutional neural networks (CNNs) or linear layers equipped with SiLU activation functions. For each modality, multiple expert modules are combined adaptively based on the learned gating outputs. 

\subsection{Overall Performance}

As presented in Table \ref{tab:detection-comparison}, our model outperforms all baselines across multiple benchmarks, surpassing DAMMFND (the strongest baseline) on both Weibo and Weibo-21 datasets. On Weibo, it improves DAMMFND’s 94.3\% F1-score and accuracy to 95.1\% (+0.7\%), and on Weibo-21 it further increases the performance from 94.7\% to 95.4\% (+0.7\%). In addition to these overall gains, our model demonstrates stronger domain-level performance: on Weibo, it outperforms DAMMFND in 6 out of 9 categories, achieving substantial improvements in Science (90.3\%, +5.2\%) and Politics (85.2\%, +3.0\%) alongside consistent gains in Education, Society, and Health; on Weibo-21, the improvements are even more pronounced, surpassing DAMMFND in 7 out of 9 categories with notable gains in Science (94.1\%, +1.8\%), Politics (98.5\%, +1.3\%), and Society (97.7\%, +1.6\%). These results suggest that our approach effectively captures domain-specific linguistic and semantic nuances, leading to more robust performance across heterogeneous misinformation types. \textbf{More results and analysis can be found in the supplementary materials.}

\subsection{Ablation Study}
We investigate the contribution of each component in our framework by constructing six ablation variants: (1) \textbf{w/o Inc}: removing the entire inconsistency–suppression branch; (2) \textbf{w/o LL Head}: discarding the low-level local reasoning head; (3) \textbf{w/o GL Head}: discarding the global-level local head; (4) \textbf{w/o GG Head}: discarding the global-level global head; (5) \textbf{w/o Hyper Modulator}: removing the hyper-network modulator that refines fused representations; and (6) \textbf{w/o Retrieval}: disabling external knowledge retrieval and relying solely on intrinsic multimodal cues.

Ablation results in Table~\ref{tab:ablation_full} demonstrate that each component contributes meaningfully to the overall performance. In particular, eliminating the inconsistency branch causes the largest degradation, with accuracy drops of \textbf{1.6\%} on Weibo and \textbf{1.7\%} on Weibo-21, underscoring the importance of explicitly harmonizing cross-modal features. Removing the low-level head yields decreases of 0.8\% and 0.9\% on the two datasets, while suppressing the global-level heads (w/o GL and w/o GG) further lowers accuracy by up to 1.3\%. The hyper modulator also proves critical: its removal degrades accuracy by 1.5\% on both datasets. Finally, turning off retrieval still harms performance (0.7\% on Weibo and 0.8\% on Weibo-21), indicating that external knowledge supplies complementary evidence beyond the learned representations.

\begin{table}[t]
  \centering
  \setlength{\tabcolsep}{3.5pt}
  % \renewcommand{\arraystretch}{1.15}
  % \tiny

  \begin{tabular}{l l c c c}
    \toprule
    \multirow{2}{*}{} & \multirow{2}{*}{\textbf{Method}}
      & \multirow{2}{*}{\textbf{Accuracy}} 
      & \multicolumn{2}{c}{\textbf{F1 score}} \\
    \cmidrule(lr){4-5}
    & & & \textbf{Real} & \textbf{Fake} \\
    \midrule
\multirow{7}{*}{\rotatebox[origin=c]{90}{\textbf{Weibo}}}
  & \textbf{Full model} 
  & \textbf{95.1} 
  & \textbf{95.2} 
  & \textbf{95.0} \\
  & \hspace{0.8em}--w/o Inconsistency branch & --1.6 & --1.6 & --1.6 \\
  & \hspace{0.8em}--w/o LL Head               & --0.8 & --0.8 & --0.9 \\
  & \hspace{0.8em}--w/o GL Head               & --1.1 & --1.1 & --1.1 \\
  & \hspace{0.8em}--w/o GG Head               & --1.3 & --1.4 & --1.3 \\
  & \hspace{0.8em}--w/o Hyper Modulator       & --1.5 & --1.5 & --1.5 \\
  & \hspace{0.8em}--w/o Retrieval             & --0.7 & --0.7 & --0.7 \\
\midrule
\multirow{7}{*}{\rotatebox[origin=c]{90}{\textbf{Weibo-21}}}
  & \textbf{Full model} 
  & \textbf{95.4} 
  & \textbf{95.5} 
  & \textbf{95.3} \\
  & \hspace{0.8em}--w/o Inconsistency branch & --1.7 & --1.7 & --1.7 \\
  & \hspace{0.8em}--w/o LL Head               & --0.9 & --0.9 & --0.9 \\
  & \hspace{0.8em}--w/o GL Head               & --1.1 & --1.1 & --1.1 \\
  & \hspace{0.8em}--w/o GG Head               & --1.3 & --1.4 & --1.4 \\
  & \hspace{0.8em}--w/o Hyper Modulator       & --1.5 & --1.5 & --1.6 \\
  & \hspace{0.8em}--w/o Retrieval             & --0.8 & --0.8 & --0.8 \\
    \bottomrule
  \end{tabular}
\caption{Ablation results showing performance (\%) for the full model, and absolute drops for ablated variants on Weibo and Weibo-21 datasets.}
\label{tab:ablation_full}
\end{table}

\subsection{Case Study}
We illustrate the framework’s interpretable results using two image-text examples. The model computes three kinds of cross modal inconsistency: LL compares words with local regions, GL compares the sentence with local regions, and GG compares the full text with the whole image. In the first example, a protest scene is captioned “lifting of all COVID-19 restrictions.” Slogans, dense crowd, and policing cues indicate a demonstration rather than a celebration, so both local and global evidence conflict with the caption; all three inconsistency scores rise and the final logit favors “fake.” In the second example, a player lifts the World Cup trophy. The trophy, jersey, and celebratory pose align with the text at both local and global levels; all three inconsistency scores remain low and the final logit favors “real.”
\begin{figure}[t]
  \centering
  \includegraphics[width=\linewidth]{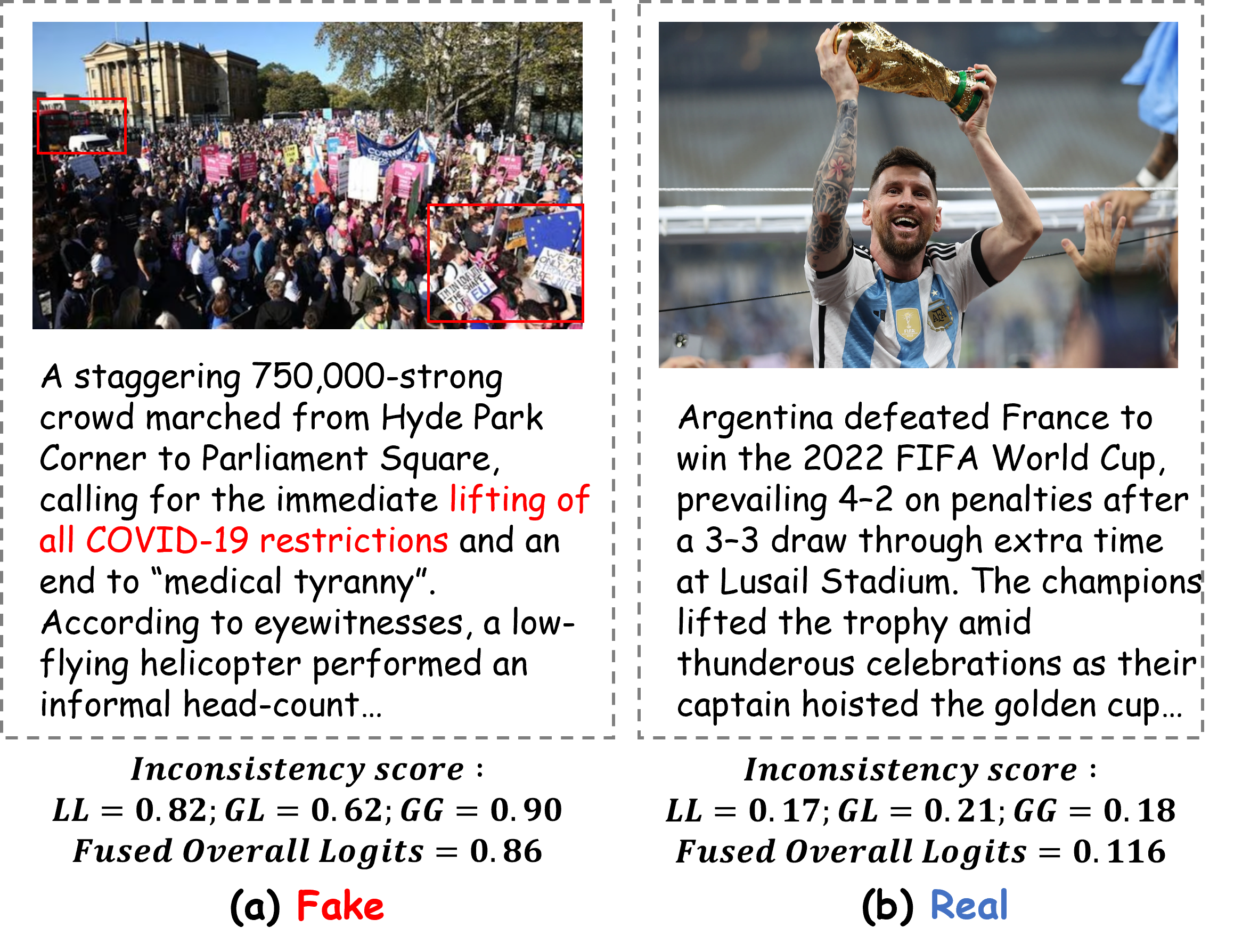}
  \caption{Case study: The left image's caption claims "lifting COVID-19 restrictions" but contradicts EU flags (red boxes), while the right image's World Cup ceremony matches its description}
  \label{fig:case_study}
\end{figure}

% \textcolor{red}{TSNE plots can also be added with description! (as the page limits of CVPR is 8 pages)}
\section{Conclusion}
We propose DMIL, a multi-granularity MIL framework for multimodal fake news detection under the collective assumption. By formulating the task in a weakly supervised setting, DMIL eliminates the need for costly, fine-grained annotations of cross-modal conflicts. The embedding-space branch aggregates multi-scale, modality-specific features via expert-guided fusion, while the instance-space branch explicitly models interactions across local-local, local-global, and global-global granularities. To further enhance cross-domain adaptability, we introduce a domain-aware reasoning module that modulates inconsistency signals using both explicit domain labels and semantically related topics. Extensive experiments across multiple datasets show that DMIL consistently outperforms strong baselines, particularly in challenging cross-domain scenarios. Our results underscore the promise of multi-granularity MIL for advancing weakly supervised multimodal learning.
{
    \small
    \bibliographystyle{ieeenat_fullname}
    \bibliography{main}

@String(CVPR= {IEEE Conf. Comput. Vis. Pattern Recog.})

@String(ECCV= {Eur. Conf. Comput. Vis.})

@String(ICME = {Int. Conf. Multimedia and Expo})

@String(ICLR = {Int. Conf. Learn. Represent.})

@String(IJCAI = {IJCAI})

@String(AAAI = {AAAI})

@String(CVPR  = {CVPR})

@String(ECCV  = {ECCV})

@String(ICME  =	{ICME})

@String(ICLR  = {ICLR})

@inproceedings{jin2017attRNN,
  author    = {Jin, Zhiwei and Cao, Juan and Guo, Han and Zhang, Yongdong and Luo, Jiebo},
  title     = {Multimodal Fusion with Recurrent Neural Networks for Rumor Detection on Microblogs},
  booktitle = {Proceedings of the 25th ACM International Conference on Multimedia},
  series    = {MM '17},
  year      = {2017},
  pages     = {795--816},
  publisher = {Association for Computing Machinery},
  address   = {New York, NY, USA},
  doi       = {10.1145/3123266.3123454},
  url       = {https://doi.org/10.1145/3123266.3123454}
}

@inproceedings{khattar2019mvae,
  author    = {Khattar, Dhruv and Goud, Jaipal Singh and Gupta, Manish and Varma, Vasudeva},
  title     = {MVAE: Multimodal Variational Autoencoder for Fake News Detection},
  booktitle = {The World Wide Web Conference},
  series    = {WWW '19},
  year      = {2019},
  pages     = {2915--2921},
  publisher = {Association for Computing Machinery},
  address   = {New York, NY, USA},
  doi       = {10.1145/3308558.3313552},
  url       = {https://doi.org/10.1145/3308558.3313552}
}

@article{singhal2020spotfake,
  author    = {Singhal, Shivangi and Kabra, Anubha and Sharma, Mohit and Shah, Rajiv Ratn and Chakraborty, Tanmoy and Kumaraguru, Ponnurangam},
  title     = {SpotFake+: A Multimodal Framework for Fake News Detection via Transfer Learning (Student Abstract)},
  journal   = {Proceedings of the AAAI Conference on Artificial Intelligence},
  volume    = {34},
  number    = {10},
  pages     = {13915--13916},
  year      = {2020},
  doi       = {10.1609/aaai.v34i10.7230},
  url       = {https://ojs.aaai.org/index.php/AAAI/article/view/7230}
}

@inproceedings{zhou2020safe,
  author    = {Zhou, Xinyi and Wu, Jindi and Zafarani, Reza},
  title     = {SAFE: Similarity-Aware Multi-modal Fake News Detection},
  booktitle = {Advances in Knowledge Discovery and Data Mining (PAKDD)},
  editor    = {Lauw, Hady W. and Wong, Raymond Chi-Wing and Ntoulas, Alexandros and Lim, Ee-Peng and Ng, See-Kiong and Pan, Sinno Jialin},
  year      = {2020},
  pages     = {354--367},
  publisher = {Springer International Publishing},
  address   = {Cham},
  doi       = {10.1007/978-3-030-47436-2_27},
  url       = {https://doi.org/10.1007/978-3-030-47436-2_27},
  isbn      = {978-3-030-47436-2}
}

@article{xue2021mcnn,
  title     = {Detecting Fake News by Exploring the Consistency of Multimodal Data},
  author    = {Junxiao Xue and Yabo Wang and Yichen Tian and Yafei Li and Lei Shi and Lin Wei},
  journal   = {Information Processing \& Management},
  volume    = {58},
  number    = {5},
  pages     = {102610},
  year      = {2021},
  issn      = {0306-4573},
  doi       = {10.1016/j.ipm.2021.102610},
  url       = {https://www.sciencedirect.com/science/article/pii/S0306457321001060}
}

@inproceedings{wang2023coolant,
  author    = {Wang, Longzheng and Zhang, Chuang and Xu, Hongbo and Xu, Yongxiu and Xu, Xiaohan and Wang, Siqi},
  title     = {Cross-modal Contrastive Learning for Multimodal Fake News Detection},
  booktitle = {Proceedings of the 31st ACM International Conference on Multimedia},
  year      = {2023},
  pages     = {5696--5704},
  publisher = {Association for Computing Machinery},
  address   = {New York, NY, USA},
  doi       = {10.1145/3581783.3613850},
  url       = {https://doi.org/10.1145/3581783.3613850}
}

@inproceedings{zhou2023clip,
  author    = {Zhou, Yangming and Yang, Yuzhou and Ying, Qichao and Qian, Zhenxing and Zhang, Xinpeng},
  title     = {Multimodal Fake News Detection via CLIP-Guided Learning},
  booktitle = {Proceedings of the 2023 IEEE International Conference on Multimedia and Expo (ICME)},
  year      = {2023},
  pages     = {2825--2830},
  doi       = {10.1109/ICME55011.2023.00480},
  publisher = {IEEE}
}

@inproceedings{wang2024msaca,
  author    = {Wang, Jiandong and Zhang, Hongguang and Liu, Chun and Yang, Xiongjun},
  title     = {Fake News Detection via Multi-scale Semantic Alignment and Cross-modal Attention},
  booktitle = {Proceedings of the 47th International ACM SIGIR Conference on Research and Development in Information Retrieval},
  series    = {SIGIR '24},
  year      = {2024},
  pages     = {2406--2410},
  publisher = {Association for Computing Machinery},
  address   = {New York, NY, USA},
  doi       = {10.1145/3626772.3657905},
  url       = {https://doi.org/10.1145/3626772.3657905}
}

@inproceedings{nan2021mdfend,
  author    = {Nan, Qiong and Cao, Juan and Zhu, Yongchun and Wang, Yanyan and Li, Jintao},
  title     = {MDFEND: Multi-domain Fake News Detection},
  booktitle = {Proceedings of the 30th ACM International Conference on Information and Knowledge Management (CIKM)},
  year      = {2021},
  pages     = {3343--3347},
  publisher = {Association for Computing Machinery},
  address   = {New York, NY, USA},
  doi       = {10.1145/3459637.3482139},
  url       = {https://doi.org/10.1145/3459637.3482139}
}

@inproceedings{silva2021cdmmfn,
  author    = {Silva, Amila and Luo, Ling and Karunasekera, Shanika and Leckie, Christopher},
  title     = {Embracing Domain Differences in Fake News: Cross-domain Fake News Detection using Multi-modal Data},
  booktitle = {Proceedings of the AAAI Conference on Artificial Intelligence},
  series    = {AAAI '21},
  year      = {2021},
  pages     = {557--565},
  publisher = {AAAI Press},
  address   = {Palo Alto, CA, USA},
  doi       = {10.1609/aaai.v35i1.16134},
  url       = {https://ojs.aaai.org/index.php/AAAI/article/view/16134}
}

@inproceedings{tong2024mmdfnd,
  author    = {Tong, Yu and Lu, Weihai and Zhao, Zhe and Lai, Song and Shi, Tong},
  title     = {MMDFND: Multi-modal Multi-Domain Fake News Detection},
  booktitle = {Proceedings of the 32nd ACM International Conference on Multimedia},
  series    = {MM '24},
  year      = {2024},
  pages     = {1178--1186},
  publisher = {Association for Computing Machinery},
  address   = {New York, NY, USA},
  doi       = {10.1145/3664647.3681317},
  url       = {https://doi.org/10.1145/3664647.3681317}
}

@inproceedings{lu2025dammfnd,
  author    = {Lu, Weihai and Tong, Yu and Ye, Zhiqiu},
  title     = {DAMMFND: Domain-Aware Multimodal Multi-view Fake News Detection},
  booktitle = {Proceedings of the AAAI Conference on Artificial Intelligence},
  series    = {AAAI '25},
  year      = {2025},
  pages     = {559--567},
  publisher = {AAAI Press},
  address   = {Palo Alto, CA, USA},
  doi       = {10.1609/aaai.v39i1.32036},
  url       = {https://ojs.aaai.org/index.php/AAAI/article/view/32036}
}

@inproceedings{devlin2019bert,
  author    = {Devlin, Jacob and Chang, Ming-Wei and Lee, Kenton and Toutanova, Kristina},
  title     = {BERT: Pre-training of Deep Bidirectional Transformers for Language Understanding},
  booktitle = {Proceedings of the 2019 Conference of the North American Chapter of the Association for Computational Linguistics: Human Language Technologies, Volume 1 (Long and Short Papers)},
  year      = {2019},
  pages     = {4171--4186},
  doi       = {10.18653/v1/N19-1423},
  publisher = {Association for Computational Linguistics}
}

@inproceedings{he2022mae,
  author    = {He, Kaiming and Chen, Xinlei and Xie, Saining and Li, Yanghao and Dollár, Piotr and Girshick, Ross},
  title     = {Masked Autoencoders Are Scalable Vision Learners},
  booktitle = {Proceedings of the 2022 IEEE/CVF Conference on Computer Vision and Pattern Recognition (CVPR)},
  year      = {2022},
  pages     = {15979--15988},
  doi       = {10.1109/CVPR52688.2022.01553},
  publisher = {IEEE}
}

@inproceedings{a2021clip,
  title={Learning Transferable Visual Models From Natural Language Supervision},
  author={Alec Radford and Jong Wook Kim and Chris Hallacy and Aditya Ramesh and Gabriel Goh and Sandhini Agarwal and Girish Sastry and Amanda Askell and Pamela Mishkin and Jack Clark and Gretchen Krueger and Ilya Sutskever},
  booktitle={International Conference on Machine Learning},
  year={2021},
  url={https://api.semanticscholar.org/CorpusID:231591445}
}

@inproceedings{wang2018eann,
  author = {Wang, Yaqing and Ma, Fenglong and Jin, Zhiwei and Yuan, Ye and Xun, Guangxu and Jha, Kishlay and Su, Lu and Gao, Jing},
  title = {EANN: Event Adversarial Neural Networks for Multi-Modal Fake News Detection},
  booktitle = {Proceedings of the 24th ACM SIGKDD International Conference on Knowledge Discovery \& Data Mining (KDD)},
  editor = {},
  pages = {849--857},
  publisher = {Association for Computing Machinery},
  volume = {},
  year = {2018},
  url = {https://doi.org/10.1145/3219819.3219903}
}

@article{shu2020fakenewsnet,
  author = {Shu, Kai and Mahudeswaran, Deepak and Wang, Suhang and Lee, Dongwon and Liu, Huan},
  title = {FakeNewsNet: A Data Repository with News Content, Social Context, and Spatiotemporal Information for Studying Fake News on Social Media},
  journal = {Big Data},
  volume = {8},
  number = {},
  pages = {171--188},
  year = {2020},
  url = {https://doi.org/10.1089/big.2020.0062},
  publisher = {Mary Ann Liebert, Inc.},
}

@article{zhou2020survey,
  author = {Zhou, Xinyi and Zafarani, Reza},
  title = {A Survey of Fake News: Fundamental Theories, Detection Methods, and Opportunities},
  journal = {ACM Comput. Surv.},
  volume = {53},
  number = {5},
  pages = {109},
  year = {2020},
  month = sep,
  publisher = {Association for Computing Machinery},
  url = {https://doi.org/10.1145/3395046},
  doi = {10.1145/3395046},
}

@article{shu2020leveraging,
  author = {Shu, Kai and Zheng, Guoqing and Li, Yichuan and Mukherjee, Subhabrata and Awadallah, Ahmed Hassan and Ruston, Scott W. and Liu, Huan},
  title = {Leveraging Multi-Source Weak Social Supervision for Early Detection of Fake News},
  journal = {CoRR},
  volume = {abs/2004.01732},
  year = {2020},
  url = {https://arxiv.org/abs/2004.01732},
  eprinttype = {arXiv},
  eprint = {2004.01732}
}

@article{shu2017fakenews,
  author = {Shu, Kai and Sliva, Amy and Wang, Suhang and Tang, Jiliang and Liu, Huan},
  title = {Fake News Detection on Social Media: A Data Mining Perspective},
  journal = {SIGKDD Explor. Newsl.},
  volume = {19},
  number = {1},
  pages = {22--36},
  year = {2017},
  month = sep,
  publisher = {Association for Computing Machinery},
  address = {New York, NY, USA},
  url = {https://doi.org/10.1145/3137597.3137600},
  doi = {10.1145/3137597.3137600}
}

@inproceedings{wu2024ueei,
  author = {Wu, Lianwei and Wang, Linyong and Zhao, Yongqiang},
  title = {Unified Evidence Enhancement Inference Framework for Fake News Detection},
  booktitle = {Proceedings of the Thirty-Third International Joint Conference on Artificial Intelligence (IJCAI)},
  year = {2024},
  pages = {723},
  url = {https://doi.org/10.24963/ijcai.2024/723},
  doi = {10.24963/ijcai.2024/723},
  location = {Jeju, Korea},
  publisher = {International Joint Conferences on Artificial Intelligence},
  isbn = {978-1-956792-04-1}
}

@article{yu2022ddchms,
  author = {Yu, En and Ma, Jianhua and Sun, Jiande and Chang, Xiaojun and Zhang, Huaxiang and Hauptmann, Alexander G.},
  title = {Deep Discrete Cross-Modal Hashing with Multiple Supervision},
  journal = {Neurocomputing},
  volume = {486},
  pages = {215--224},
  year = {2022},
  issn = {0925-2312},
  doi = {10.1016/j.neucom.2021.11.035},
  url = {https://www.sciencedirect.com/science/article/pii/S0925231221017124}
}

@inproceedings{tan2020inconsistency,
  author = {Tan, Reuben and Plummer, Bryan and Saenko, Kate},
  title = {Detecting Cross-Modal Inconsistency to Defend Against Neural Fake News},
  booktitle = {Proceedings of the 2020 Conference on Empirical Methods in Natural Language Processing (EMNLP)},
  year = {2020},
  month = nov,
  pages = {2081--2106},
  publisher = {Association for Computational Linguistics},
  address = {Online},
  url = {https://aclanthology.org/2020.emnlp-main.163/},
  doi = {10.18653/v1/2020.emnlp-main.163}
}

@article{wu2023mfir,
  author = {Wu, Lianwei and Long, Yuzhou and Gao, Chao and Wang, Zhen and Zhang, Yanning},
  title = {MFIR: Multimodal Fusion and Inconsistency Reasoning for Explainable Fake News Detection},
  journal = {Information Fusion},
  volume = {100},
  pages = {101944},
  year = {2023},
  issn = {1566-2535},
  doi = {10.1016/j.inffus.2023.101944},
  url = {https://www.sciencedirect.com/science/article/pii/S1566253523002609}
}

@article{xiong2023trimoon,
  author = {Xiong, Shufeng and Zhang, Guipei and Batra, Vishwash and Xi, Lei and Shi, Lei and Liu, Liangliang},
  title = {TRIMOON: Two-Round Inconsistency-based Multi-modal Fusion Network for Fake News Detection},
  journal = {Information Fusion},
  volume = {93},
  pages = {150--158},
  year = {2023},
  issn = {1566-2535},
  doi = {10.1016/j.inffus.2022.12.016},
  url = {https://www.sciencedirect.com/science/article/pii/S1566253522002639}
}

@article{guo2025crossmodal,
  author = {Guo, Ying and Li, Bingxin and Zhen, Kexin and Liu, Jie and Li, Gaolei and Wang, Qi and Liu, Yong-Jin},
  title = {Consistency-Heterogeneity Balanced Fake News Detection via Cross-Modal Matching},
  journal = {IEEE Transactions on Artificial Intelligence},
  volume = {6},
  number = {07},
  pages = {1787--1796},
  year = {2025},
  month = jul,
  publisher = {IEEE Computer Society},
  address = {Los Alamitos, CA, USA},
  issn = {2691-4581},
  doi = {10.1109/TAI.2025.3527921},
  url = {https://doi.ieeecomputersociety.org/10.1109/TAI.2025.3527921}
}

@article{carbonneau2018multiple,
  title={Multiple instance learning: A survey of problem characteristics and applications},
  author={Carbonneau, Marc-Andr{\'e} and Cheplygina, Veronika and Granger, Eric and Gagnon, Ghyslain},
  journal={Pattern recognition},
  volume={77},
  pages={329--353},
  year={2018},
  publisher={Elsevier}
}

@article{chen2023altclip,
  author = {Chen, Zhongzhi and Liu, Guang and Zhang, Bo-Wen and Yang, Qinghong and Wu, Ledell},
  title = {AltCLIP: Altering the Language Encoder in CLIP for Extended Language Capabilities},
  journal = {Findings of the Association for Computational Linguistics: ACL 2023},
  volume = {2023},
  number = {Jul},
  pages = {8666--8682},
  year = {2023},
  month = jul,
  publisher = {Association for Computational Linguistics},
  address = {Toronto, Canada},
  doi = {10.18653/v1/2023.findings-acl.552},
  url = {https://aclanthology.org/2023.findings-acl.552/}
}

@article{maron1997framework,
  title={A framework for multiple-instance learning},
  author={Maron, Oded and Lozano-P{\'e}rez, Tom{\'a}s},
  journal={Advances in neural information processing systems},
  volume={10},
  year={1997}
}

@article{foulds2010review,
  title={A review of multi-instance learning assumptions},
  author={Foulds, James and Frank, Eibe},
  journal={The knowledge engineering review},
  volume={25},
  number={1},
  pages={1--25},
  year={2010},
  publisher={Cambridge University Press}
}

@incollection{herrera2016multiple,
  title={Multiple instance learning},
  author={Herrera, Francisco and Ventura, Sebasti{\'a}n and Bello, Rafael and Cornelis, Chris and Zafra, Amelia and S{\'a}nchez-Tarrag{\'o}, D{\'a}nel and Vluymans, Sarah},
  booktitle={Multiple instance learning: foundations and algorithms},
  pages={17--33},
  year={2016},
  publisher={Springer}
}

@article{quellec2017multiple,
  title={Multiple-instance learning for medical image and video analysis},
  author={Quellec, Gwenol{\'e} and Cazuguel, Guy and Cochener, B{\'e}atrice and Lamard, Mathieu},
  journal={IEEE reviews in biomedical engineering},
  volume={10},
  pages={213--234},
  year={2017},
  publisher={IEEE}
}

@inproceedings{ma2018mmoe,
  author    = {Ma, Jiaqi and Zhao, Zhe and Yi, Xinyang and Chen, Jilin and Hong, Lichan and Chi, Ed H.},
  title     = {Modeling Task Relationships in Multi-task Learning with Multi-gate Mixture-of-Experts},
  booktitle = {Proceedings of the 24th ACM SIGKDD International Conference on Knowledge Discovery \& Data Mining},
  series    = {KDD '18},
  year      = {2018},
  pages     = {1930--1939},
  publisher = {Association for Computing Machinery},
  address   = {New York, NY, USA},
  doi       = {10.1145/3219819.3220007},
  url       = {https://doi.org/10.1145/3219819.3220007}
}

@inproceedings{qin2020mose,
  author    = {Qin, Zhen and Cheng, Yicheng and Zhao, Zhe and Chen, Zhe and Metzler, Donald and Qin, Jingzheng},
  title     = {Multitask Mixture of Sequential Experts for User Activity Streams},
  booktitle = {Proceedings of the 26th ACM SIGKDD International Conference on Knowledge Discovery \& Data Mining},
  series    = {KDD '20},
  year      = {2020},
  pages     = {3083--3091},
  publisher = {Association for Computing Machinery},
  address   = {New York, NY, USA},
  doi       = {10.1145/3394486.3403359},
  url       = {https://doi.org/10.1145/3394486.3403359}
}

@article{zhu2023m3fend,
  author    = {Zhu, Yongchun and Sheng, Qiang and Cao, Juan and Nan, Qiong and Shu, Kai and Wu, Minghui and Wang, Jindong and Zhuang, Fuzhen},
  title     = {Memory-Guided Multi-View Multi-Domain Fake News Detection},
  journal   = {IEEE Transactions on Knowledge and Data Engineering},
  volume    = {35},
  number    = {7},
  pages     = {7178--7191},
  year      = {2023},
  publisher = {IEEE Educational Activities Department},
  address   = {USA},
  doi       = {10.1109/TKDE.2022.3185151},
  url       = {https://doi.org/10.1109/TKDE.2022.3185151}
}

@inproceedings{zhu2025emosym,
  title={Emosym: A symbiotic framework for unified emotional understanding and generation via latent reasoning},
  author={Zhu, Yijie and Lyu, Yibo and Yu, Zitong and Shao, Rui and Zhou, Kaiyang and Nie, Liqiang},
  booktitle={Proceedings of the 33rd ACM International Conference on Multimedia},
  pages={5451--5460},
  year={2025}
}

@inproceedings{zhu2026h,
  title={H-gar: A hierarchical interaction framework via goal-driven observation-action refinement for robotic manipulation},
  author={Zhu, Yijie and Shao, Rui and Liu, Ziyang and He, Jie and Liu, Jizhihui and Wang, Jiuru and Yu, Zitong},
  booktitle={Proceedings of the AAAI Conference on Artificial Intelligence},
  volume={40},
  number={22},
  pages={18882--18890},
  year={2026}
}

@article{zhu2026uniemo,
  title={Uniemo: Unifying emotional understanding and generation with learnable expert queries},
  author={Zhu, Yijie and Zhang, Lingsen and Yu, Zitong and Shao, Rui and Tan, Tao and Nie, Liqiang},
  journal={IEEE Transactions on Image Processing},
  year={2026},
  publisher={IEEE}
}

@article{zhu2026delta,
  title={{$\Delta$-VLA}: Prior-Guided Vision-Language-Action Models via World Knowledge Variation},
  author={Zhu, Yijie and He, Jie and Shao, Rui and Yuan, Kaishen and Tan, Tao and Yuan, Xiaochen and Yu, Zitong},
  journal={arXiv preprint arXiv:2603.08361},
  year={2026}
}

@article{zhu2026ati,
  title={ATI-VLA: Action-Centric Predictive Vision-Language-Action Models via Actionable Alignment Then Adaptive Injection},
  author={Zhu, Yijie and Shao, Rui and He, Jie and Li, Wei and Zhao, Bo and Wang, Yelin and Yuan, Xiaochen and Tan, Tao and Zhang, Miao and Peng, Xiaojiang and others},
  journal={arXiv preprint arXiv:2610.01741},
  year={2026}
}

@article{zhu2026learning,
  title={Learning to Use Imagination: Progress-Conditioned Future Utilization for World Action Models},
  author={Zhu, Yijie and Yu, Zitong and Li, Wei and Ma, Hui and Li, Wen and Shao, Rui and Nie, Liqiang},
  journal={arXiv preprint arXiv:2609.06578},
  year={2026}
}

@article{wang2026affectagent,
  title={AffectAgent: Collaborative Multi-Agent Reasoning for Retrieval-Augmented Multimodal Emotion Recognition},
  author={Wang, Zeheng and Yu, Zitong and Zhu, Yijie and Zhao, Bo and Liang, Haochen and Wang, Taorui and Xia, Wei and Zhang, Jiayu and Liu, Zhishu and Ma, Hui and others},
  journal={arXiv preprint arXiv:2604.12735},
  year={2026}
}

@article{wang2026navigating,
  title={Navigating the Emotion Tree: Hierarchical Hyperbolic RAG for Multimodal Emotion Recognition},
  author={Wang, Zeheng and Zhao, Bo and Zhu, Yijie and Liu, Zhishu and Ma, Hui and Zhang, Ruixin and Ding, Shouhong and Xie, Qianyu and Yu, Zitong},
  journal={arXiv preprint arXiv:2605.18884},
  year={2026}
}

@misc{gao2026polarityasymmetricstructuralcalibrationlink,
  title={Polarity-Asymmetric Structural Calibration for Link Sign Prediction},
  author={Qiqi Gao and Wenzhuo Song and Xueyan Liu},
  year={2026},
  eprint={2609.05896},
  archivePrefix={arXiv},
  primaryClass={cs.SI},
  url={https://arxiv.org/abs/2609.05896}
}

@misc{wang2026signedgraph,
  author        = {Wang, Zifan and Chen, Siyu and Song, Wenzhuo},
  title         = {Signed-Graph Recommendation as Structural Consistency Maximization},
  year          = {2026},
  eprint        = {2607.05952},
  archivePrefix = {arXiv},
  primaryClass  = {cs.SI},
  doi           = {10.48550/arXiv.2607.05952}
}

@inproceedings{wang2027manchutts,
  author    = {Wang, Suhua and Wang, Zifan and Sun, Xiaoxin and Wang, Meiqi and Zhang, Yun and Li, Xin},
  title     = {{ManchuTTS}: Towards High-Quality Manchu Speech Synthesis via Flow Matching and Hierarchical Text Representation},
  booktitle = {Advanced Intelligent Computing Technology and Applications},
  editor    = {Huang, De-Shuang and Li, Bo and Zhang, Qinhu and Bao, Wenzheng},
  series    = {Lecture Notes in Computer Science},
  volume    = {16669},
  pages     = {51--61},
  publisher = {Springer Nature Singapore},
  address   = {Singapore},
  year      = {2027},
  doi       = {10.1007/978-981-92-3492-9_5}
}

@article{ma2026cm,
  title={CM-CLIP: few-shot cross-modal generalization for face anti-spoofing via prompt learning},
  author={Ma, Hui and Liu, Ajian and Lin, Xun and Escalante, Hugo Jair and Guyon, Isabelle and Wan, Jun and Lei, Zhen and Liang, Yanyan},
  journal={Machine Intelligence Research},
  volume={23},
  number={2},
  pages={366--382},
  year={2026},
  publisher={Springer}
}

@article{ma2026clip,
  title={CLIP-SA: CLIP-Guided Semantic Alignment for Generalizable Face Anti-spoofing},
  author={Ma, Hui and Liu, Ajian and Xue, Shuai and Diao, Yunfeng and Liang, Dashuang and Liang, Hailun and Liang, Yanyan and Wan, Jun and Yang, Xun and Hu, ZhenZhen and others},
  journal={IEEE Transactions on Multimedia},
  year={2026},
  publisher={IEEE}
}

@article{ma2026flexible,
  title={Flexible Modal Mixture-of-Experts with Inter-Modal Knowledge Distillation for Face Anti-Spoofing},
  author={Ma, Hui and Liu, Ajian and Li, Ning and Yang, Liying and Wang, Boyun and Zou, Hang and Zhang, Yuan and Huang, Jing and Pu, Zhiqiang and Wan, Jun and others},
  journal={IEEE Transactions on Information Forensics and Security},
  year={2026},
  publisher={IEEE}
}

@article{liu2026aullm++,
  title={AULLM++: Structural Reasoning with Large Language Models for Micro-Expression Recognition},
  author={Liu, Zhishu and Yuan, Kaishen and Zhao, Bo and Ma, Hui and Yu, Zitong},
  journal={arXiv preprint arXiv:2603.08387},
  year={2026}
}

@article{wang2026micro,
  title={Micro-gesture recognition: A comprehensive survey of datasets, methods, and challenges},
  author={Wang, Taorui and Lin, Xun and Xu, Yong and Ye, Qilang and Guo, Dan and Escalera, Sergio and Khoriba, Ghada and Yu, Zitong},
  journal={Machine Intelligence Research},
  volume={23},
  number={2},
  pages={308--330},
  year={2026},
  publisher={Springer}
}

@inproceedings{10.1145/3767308.3835254,
  author = {Wang, Taorui and Xia, Wei and Ma, Hui and Song, Zijia and Zhang, Jiayu and Wang, Zeheng and Xu, Yong and Yu, Zitong},
  title = {GMoT: Gated Motion-Aware Tokenization for Fine-Grained Micro-Gesture Video Reasoning with Multimodal LLMs},
  year = {2026},
  isbn = {9798400722134},
  publisher = {Association for Computing Machinery},
  address = {New York, NY, USA},
  url = {https://doi.org/10.1145/3767308.3835254},
  doi = {10.1145/3767308.3835254},
  booktitle = {Proceedings of the 34th ACM International Conference on Multimedia},
  pages = {10596--10605},
  numpages = {10},
  location = {Rio de Janeiro, Brazil},
  series = {MM '26}
}

@inproceedings{10.1007/978-981-92-4529-1_3,
  author = {Wang, Taorui and Ma, Hui and Xia, Wei and Zhang, Jiayu and Huang, Jiajian and Xu, Yong},
  editor = {Chen, Haoyu and Ren, Zhao and Wang, Jiadong and Provost, Emily and Adeli, Ehsan and Schuller, Bj{\"o}rn W. and Schultz, Tanja and Zhao, Guoying},
  title = {SwinT-Enhanced Class-Wise Fusion for Micro-gesture Classification},
  booktitle = {Human Behavior Analysis and Interaction for Emotional Intelligence},
  year = {2027},
  publisher = {Springer Nature Singapore},
  address = {Singapore},
  pages = {28--34},
  isbn = {978-981-92-4529-1}
}

@article{11658652,
  author={Wang, Yelin and Zhu, Yijie and Song, Zijia and Ma, Hui and Wang, Zeheng and Yu, Zitong and Ma, Fei and Tian, Qi},
  journal={IEEE Transactions on Affective Computing},
  title={Uni-EmoAgent: A Unified Multi-Agent Framework for Affective Understanding and Art-Oriented Emotional Image Generation},
  year={2026},
  pages={1--13},
  doi={10.1109/TAFFC.2026.3725184}
}

@inproceedings{wang2026rsiccllm,
  title     = {{RSICCLLM}: A Multimodal Large Language Model for Remote Sensing Image Change Captioning},
  author    = {Wang, Yelin and Song, Zijia and Ye, Shuo and Yang, Chuanguang and Wang, Miaoyu and Xu, Yong and An, Zhulin and Xu, Yongjun and Yu, Zitong},
  booktitle = {Computer Vision -- ECCV 2026},
  series    = {Lecture Notes in Computer Science},
  volume    = {17056},
  pages     = {253--272},
  publisher = {Springer},
  year      = {2026},
  doi       = {10.1007/978-3-032-37324-3_14}
}

@inproceedings{wang2026dfm,
  title     = {{DFM}: Difference Feature Modeling with Text-Guided Gated Contrastive Loss for Remote Sensing Image Change Captioning},
  author    = {Wang, Yelin and Song, Zijia and Yang, Chuanguang and Wang, Miaoyu and An, Zhulin and Huang, Libo and Xu, Yongjun},
  booktitle = {2026 IEEE International Conference on Multimedia and Expo (ICME)},
  year      = {2026},
  url       = {https://arxiv.org/abs/2606.27410}
}

@inproceedings{tan2026veritas,
  title={Veritas: Generalizable deepfake detection via pattern-aware reasoning},
  author={Tan, Hao and Tan, Zichang and Shi, Senyuan and Liu, Ajian and Song, Chuanbiao and Zhu, Huijia and Wang, Weiqiang and Wan, Jun and Lei, Zhen and others},
  booktitle={International Conference on Learning Representations},
  volume={2026},
  pages={66420--66477},
  year={2026}
}

@article{tan2026videoveritas,
  title={Videoveritas: Ai-generated video detection via perception pretext reinforcement learning},
  author={Tan, Hao and Lan, Jun and Shi, Senyuan and Tan, Zichang and Yu, Zijian and Zhu, Huijia and Wang, Weiqiang and Wan, Jun and Lei, Zhen},
  journal={arXiv preprint arXiv:2602.08828},
  year={2026}
}

@article{shi2026hydraprompt,
  title={HydraPrompt: An Adaptive and Asymmetric Framework of Vision-Language Models for Synthetic Image Detection},
  author={Shi, Senyuan and Tan, Hao and Tan, Zichang and Feng, Shuhan and Liu, Ajian and Escalera, Sergio and Wan, Jun},
  journal={arXiv preprint arXiv:2605.26421},
  year={2026}
}

@inproceedings{DBLP:conf/iclr/PanCWDW26,
  author       = {Yu Pan and Jiahao Chen and Lin Wang and Bingrong Dai and Wenjie Wang},
  editor       = {Carl Vondrick and Bharath Hariharan and Colin Raffel and Lerrel Pinto and Diyi Yang and Aleksandra Faust},
  title        = {STEDiff: Revealing the Spatial and Temporal Redundancy of Backdoor Attacks in Text-to-Image Diffusion Models},
  booktitle    = {The Fourteenth International Conference on Learning Representations, {ICLR} 2026, Rio de Janeiro, Brazil, April 23-27, 2026},
  publisher    = {proceedings.iclr.cc},
  year         = {2026},
  url          = {https://proceedings.iclr.cc/paper\_files/paper/2026/hash/559726fdfb19005e368be4ce3d40e3e5-Abstract-Conference.html}
}

@article{DBLP:journals/corr/abs-2605-07399,
  author       = {Yu Pan and Andi Zhang and Yi Wang and Sibei Yang and Wenjie Wang},
  title        = {{GPO-V:} Jailbreak Diffusion Vision Language Model by Global Probability Optimization},
  journal      = {CoRR},
  volume       = {abs/2605.07399},
  year         = {2026},
  url          = {https://doi.org/10.48550/arXiv.2605.07399},
  doi          = {10.48550/ARXIV.2605.07399},
  eprinttype   = {arXiv},
  eprint       = {2605.07399}
}
}

% WARNING: do not forget to delete the supplementary pages from your submission 
% \input{sec/X_suppl}

\end{document}